\documentclass{article}

\usepackage{amsfonts}%
\usepackage{amsmath}%
\usepackage{amssymb}%
\usepackage{amsthm}%
\usepackage{booktabs}
\usepackage{bm}

\usepackage{graphicx} % more modern
\usepackage{subfigure} 

\usepackage{natbib}

\usepackage{algorithm}
\usepackage{algorithmic}

\usepackage{hyperref}

\usepackage[accepted]{icml2011}

\icmltitlerunning{Conditional Anomaly Detection Using Soft Harmonic Functions:
An Application to Clinical Alerting}	

\newcommand{\commentout}[1]{}
\newcommand{\eat}[1]{}

\newcommand{\bx}{{\bf x}}

\newcommand{\by}{{\bf y}}

\newcommand{\bell}{{\bm \ell}}

\newcommand{\realset}{\mathbb{R}}

\newcommand{\transpose}{^\mathsf{\scriptscriptstyle T}}

\begin{document} 
\twocolumn[
\icmltitle{Conditional Anomaly Detection Using Soft Harmonic Functions:\\
An Application to Clinical Alerting}
% It is OKAY to include author information, even for blind
% submissions: the style file will automatically remove it for you
% unless you've provided the [accepted] option to the icml2011
% package.
\vskip -0.5cm
\icmlauthor{Michal Valko}{michal@cs.pitt.edu}%\icmladdress{Computer Science Department, University of Pittsburgh, PA} 
\vskip -0.05cm
\icmladdress{Computer Science Department, University of Pittsburgh, PA}
\vskip -0.05cm
\icmlauthor{Hamed Valizadegan}{hamed@cs.pitt.edu}
\vskip -0.05cm
\icmladdress{Computer Science Department, University of Pittsburgh, PA}
\vskip -0.05cm
\icmlauthor{Branislav Kveton}{branislav.kveton@technicolor.com}
\vskip -0.05cm
\icmladdress{Technicolor, Palo Alto, PA} 
\vskip -0.05cm
\icmlauthor{Gregory F.~Cooper}{gfc@pitt.edu}
\vskip -0.05cm
\icmladdress{Department of Biomedical Informatics, University of Pittsburgh, PA}
\vskip -0.05cm
\icmlauthor{Milos Hauskrecht}{milos@cs.pitt.edu}
\vskip -0.05cm
\icmladdress{Computer Science Department, University of Pittsburgh, PA}
\vskip -0.1cm
%\vskip -0.05cm
%\icmladdress{Computer Science Department, University of Pittsburgh, PA}
% You may provide any keywords that you 
% find helpful for describing your paper; these are used to populate 
% the "keywords" metadata in the PDF but will not be shown in the document
\icmlkeywords{label propagation, anomaly detection, graph based learning, health-care}
\vskip 0.2in
]

\begin{abstract} 
%Most current work on anomaly detection attempts to identify unusual data instances that deviate from the rest of the data.
Timely detection of concerning events is an important problem in clinical practice. 
In this paper, we consider the problem of conditional anomaly detection that aims to identify data instances with an unusual response, 
such as the omission of an important lab test. We develop a new non-parametric approach for conditional anomaly detection based on the soft harmonic solution, with which we estimate the confidence of the label to detect anomalous mislabeling. We further regularize the solution to avoid the detection of isolated examples and examples on the boundary of the distribution support. We demonstrate the efficacy of the proposed method in detecting unusual labels on a real-world electronic health record dataset and compare it to several baseline approaches.

	\eat{We also investigate the online formulation of the problem and make use of graph quantization to
	deliver a scalable algorithm.}% 
\end{abstract} 

\section{Introduction}
\label{sec:Introduction}

\hyphenation{ana-ly-sis}

With the advances in health-care and with more data being handled and stored electronically, the opportunities 
increase for machine learning to improve the health care. 
Despite continuous improvement in medical practice, medical errors remain a very serious problem.
According to recent data, medical errors are the 8-th leading cause of death in the US population \cite{kohn2000to}.
We aim to identify medical errors that correspond to unusual patient-management decisions, such as ordering a medication.
Our hypothesis is that patient-management decisions that are unusual with respect to past patients may be due to
errors and that it is worthwhile to raise an alert if such a condition is encountered.
Typical systems for medical error detection rely on the clinical knowledge, such as expert-derived rules. 
Extracting such knowledge is costly, time-consuming, and very difficult with medical practice constantly changing. Machine learning can offer a  
viable alternative: using past medical records to detect anomalies. 
Detecting anomalies in the patient-management decisions
has the potential to help avoid medical errors, which could lead to
improved quality of care and decreased costs.  

\commentout{ (Milos)
As the health-care data become handled and stored electronically, the opportunities
rise for machine learning to improve the health care.
Despite continuous improvement in medical practice, medical errors remain a very serious problem.
We aim to identify such medical errors that correspond to unusual patient-management decisions, such as ordering a medication.
Our hypothesis is that patient-management decisions that are unusual with respect to past patients may be due to
errors and that it is worthwhile to raise an alert if such a condition is encountered.
We seek for such patient-management decisions which are \textit{conditionally}
anomalous with respect to patient's conditions.
} %%% end commentout (Milos)

Traditional anomaly detection methods used in data analysis are unconditional and look for outliers with respect to all data attributes
\cite{markou2003novelty}.
The conditional anomaly detection (CAD) problem \cite{hauskrecht2007evidence-based,song2007conditional} seeks to detect unusual values for a subset of variables given the values of the remaining variables. 
In the special case when the target variable is a class label, the problem is often called mislabeling detection. While we focus on this special case of anomalies in the class label, the main objective is different: we are interested in constructing a system that can raise an alert when an anomalous label of a new example is observed. 
Formally, we want to solve the following problem:

%\begin{quotation}
%\noindent {\bf Problem statement} ($\bigstar$): For a dataset $(\bx_i,y_i)_{i=1}^n$
%find pairs of $(\bx_i,y_i)$ such that the probability of a different label  $P(y\ne y_i|\bx_i)$ is high.
%\end{quotation}

\begin{quotation}
\noindent {\bf Problem statement} ($\bigstar$): Given a set of $n$ past observed examples $(\bx_i,y_i)_{i=1}^{n}$ (with possible label noise), check if any instance $i$ in  recent $m$ examples $(\bx_i,y_i)_{i=n+1}^{n+m}$ is unusual. 
\end{quotation}

\noindent In general, we seek to reliably identify anomalies on the response (decision or class) variables for all possible values of the context (input or feature) variables.
Not knowing the underlying model, that generates the (attributes, label) pairs, may lead to two major complications. 
First, a given instance may be far from the past observed data points (e.g. patient cases). Because of the lack of the support for alternative responses, it is difficult to asses the anomalousness of these instances.  We refer to these instances as \emph{isolated points}. Second, the examples on the boundary of the class distribution support, also known as \emph{fringe points}, may look anomalous due to their low likelihood. %These boundary examples are also known as \emph{fringe points}.
%These boundary examples are known as \emph{fringe points} \cite{papadimitriou2003cross-outlier}.

Because the underlying conditional distribution of the data is unknown, a non-parametric approach that looks for the label consistency of the instances on their neighborhood (e.g. $k$-nearest neighbor or $k$-NN) can be very useful \cite{papadimitriou2003cross-outlier}. 
The problem with relying on models such as $k$-NN is that they fail to detect clusters of anomalous instances. 
Our approach differs from typical local neighborhood approaches in two important aspects.
First, it respects the structure of the manifold and accounts for more complex interactions in the data. Second, it solves the problem of isolated and fringe points by decreasing the confidence in predicting an opposite label for such points through regularization.

\section{Background}
\label{sec:Background}

Label propagation on the graph is widely used for semi-supervised learning (SSL).
The general idea is to assume the consistency of labels among the data which are
1) close to each other and 2) lie on the structure (manifold/cluster).
The two examples are the \emph{Consistency Method} of 
Zhou~et~al.~\cite{zhou2004learning}
and the \emph{Harmonic Solution} of 
Zhu~et~al.~\cite{zhu2003semi-supervised},
%We build on the harmonic solution \cite{zhu2003semi-supervised} and show how to regularize it
%such that we can avoid the detection of unconditional outliers and fringe points.
\eat{
The inference of the labels $\{y_i\}_{i=1}^n$ by the approach of 
\citeauthor{zhu2003semi-supervised} \shortcite{zhu2003semi-supervised}
can be interpreted as a random walk on  $G$ with the transition matrix $P = D^{-1}W$.
The harmonic solution satisfies the \emph{harmonic} property $\ell_i = \frac{1}{d_i}
\sum_{j \sim i} w_{ij} \ell_j$\footnote{$j$ and $i$ are neighbors in  $G$}.%, hence the name harmonic solution.
}
%Harmonic solution and consistency method
%Both of them 
%are the instances of a bigger class of the optimization problems called
which are the instances of 
unconstrained regularization \cite{cortes2008stability}. 
%\subsection{Notation}% With Unconstrained Regularization}
Let $G$ be the similarity graph with the nodes corresponding to $\{\bx_i\}_{i=1}^{n+m}$ and with the weighted edges $W$
encoding pairwise similarities between the nodes.
We denote by $\mathcal{L}(W)$ the (unnormalized) graph Laplacian defined as $\mathcal{L}(W)= D - W$
where $D$ is a diagonal matrix whose entries are given by $d_{ii} = \sum_j w_{ij}$.
In the transductive setting, the unconstrained regularization searches for soft (continuous) label assignment such that it maximizes fit to the labeled data and penalizes for not following the manifold structure:

\begin{equation}
  \bell^\star = \min_{\bell \in \realset^n} \
  (\bell - \by)\transpose C (\bell - \by) + \bell\transpose K \bell,
	\label{eq:unconstrained regularization}
\end{equation}

\noindent
where $K$ is a symmetric regularization matrix and $C$ is a symmetric matrix of empirical weights.
$C$ is  usually diagonal and the diagonal entries often equal to some fixed constant $c_l$ for
the labeled data and $c_u$ for the unlabeled. 
In a SSL setting,
$\by$ is a vector of pseudo-targets such that $y_i\in\{\pm1\}$ is the label of the $i$-th example when the example is labeled,
and $y_i = 0$ otherwise. 
The appealing property of \eqref{eq:unconstrained regularization} is that it can be computed by the following closed form solution:

\begin{equation}
\bell^\star  = (C^{-1}K+I)^{-1}\by
\label{eq:closed form}
\end{equation}

\eat{\cite{valko2010online} $K = L + \gamma_g I$
For regularized, is the regularized Laplacian of the
similarity graph.
}

\section{Methodology}
\label{sec:Methods}

%\subsection{Conditional Anomaly Detection}
%\label{sec:CAD}

\noindent We now propose a way to compute the anomaly score from~\eqref{eq:closed form}.
The output $\ell^\star$ of \eqref{eq:unconstrained regularization} for the example $i$ can be rewritten 
($\mathrm{sgn}$ refers to the sign function) as:
\begin{equation}
\ell_i^\star = |\ell_i^\star| \times \mathrm{sgn}(\ell_i^\star)
\label{eq:absform}
\end{equation}
\noindent SSL methods use $\mathrm{sgn}(\ell^\star_i)$ in  \eqref{eq:absform}
as the predicted label for $i$. For an unlabeled example, the closer the value of $\ell_i$ is to $\pm 1$,
the more consistent labeling information was propagated to it.
%Typically, that means that the example is close to the labeled examples of the respective class.
The key observation, which we exploit in this paper, is
that we can interpret $|\ell^\star_i|$ as a confidence of the label.
%Our situation differs from SSL becasuse we aim to asses the confidence of \emph{already labeled} example.
%Therefore, 
We define the \emph{anomaly score} $s_i$ as the
absolute difference between the actual label $y_i$ and the inferred soft label $\ell_i$:
%However, we can still take advantage of treating $|\bell^\star|$ as a confidence
%as we expect mislabeled data to be ``surrounded"  by the data with different labels.

\begin{equation}
s_i = |\ell_i^\star - y_i|.
\label{eq:anomalyscore}
\end{equation}

%Intuitively in SSL setting, the data points closer to the labeled example would have higher confidence.
%\subsection{Regularized Harmonic Solution for CAD}
%\label{sec:reg HFS}
\noindent We will now address the problems of isolated examples by setting $K = \mathcal{L}(W) + \gamma_gI$, where we diagonally regularize the graph Laplacian.
Intuitively, such a regularization lowers the confidence value $|\ell_i^\star|$ of all examples; however it reduces the confidence score of outlier points relatively more. 
In the fully labeled setting, the \textit{hard} harmonic solution \cite{zhu2003semi-supervised} degenerates to the weighted $k$-NN.
To alleviate this problem, we allow labels to spread on the graph by
using soft constraints in the unconstrained regularization problem \eqref{eq:unconstrained regularization}.
In particular, instead of $c_l=\infty$ we set $c_l$ to a finite constant and we set $C = c_l I$.
With such a setting we can solve \eqref{eq:unconstrained regularization}
using \eqref{eq:closed form}:
\begin{align}
\bell^\star %%& = \left(\left(c_l I\right)^{-1}\left(\mathcal{L}(W)+\gamma_g\right) + I\right)^{-1}\by \\
      & = \left(c_l^{-1}\mathcal{L}(W)+\left(1+\frac{\gamma_g}{c_l}\right)I\right)^{-1}\by.
			\label{eq:had}
\end{align}
\noindent To avoid computation of the inverse, we may calculate \eqref{eq:had} by solving a system of linear equations.
%\begin{equation}
%\left(c_l^{-1}\mathcal{L}(W)+\left(1+\frac{\gamma_g}{c_l}\right)I\right) \bell^\star = \by
%\label{eq:had:sle}
%\end{equation}
\eat{
[Ie. $\left(c_l^{-1}\mathcal{L}(W)+\left(1+\frac{\gamma_g}{c_l}\right)I\right) \backslash \by$ with \textsc{Matlab}]
}
%\noindent
%Since $\bell$ was not constrained to be either +1 or -1  we can interpret $|\bell|$ as a confidence of the labeling.
We then plug the output of \eqref{eq:had} into \eqref{eq:anomalyscore} to get the anomaly score.
We will refer to this score as SoftHAD score.
Intuitively, when the confidence is high but $\mathrm{sgn}(\ell_i^\star) \ne y_i$, we will consider the label $y_i$ of the case
$(\bx_i, y_i)$  anomalous.

\paragraph{Backbone Graph}
\label{sec:BackboneGraph}

The computation of the system of linear equations \eqref{eq:had}
scales with cubic\footnote{The complexity can be further improved to $O(n_u^{2.376})$ with
the Coppersmith-Winograd algorithm.} time complexity.
This is not feasible for a graph with more than several thousands of nodes.
To address the problem, we use \emph{data quantization} \cite{gray1998quantization}
and sample a set of nodes from the training data to create $G$.
We then substitute the nodes in the graph with a smaller set of $k \ll n + m$ distinct centroids
which results in $O(k^3)$.

\section{Experiments}
\label{sec:Experiments}

To evaluate  our SoftHAD method, we compare  it to the following baselines:
%\begin{iteze}
%\item 
 (1) 1-class SVM approach in which we cover each class by a separate 1-class SVM \cite{scholkopf1999estimating},
%with RBF kernel and the anomaly score equals to the distance of the example from the learned boundary of its own class. 
%This method is an example of the traditional anomaly detection method adopted for CAD.
%\item 
 (2) Quadratic discriminant analysis (QDA) model \cite{hastie2001elements}, %, where we model each class by a multivariate Gaussian,
%and the anomaly score is the class posterior of the opposite class.
%\item
 (3) SVM classification model \cite{vapnik1995nature} with RBF kernel, %where we consider an example anomalous if it falls far on
%the opposite side of the decision boundary. 
%This method is an example of the classification method adopted for CAD and
%was used  by 
%\citeauthor{valko2008conditional} \shortcite{valko2008conditional}.
 and (4)
%\item 
Weighted $k$-NN approach \cite{hastie2001elements} that uses the same weight metric $W$.% as SoftHAD
%., but relies on only on the labels in the local neighborhood and does not account for the manifold structure.
%\end{itemize}

\subsection{UCI ML Datasets}
\label{sec:UCIMLDatasets}

We first evaluated our method on the three UCI ML datasets~\cite{asuncion2007uci} for 
which an ordinal response variable was available to calculate the true anomaly score. In particular, we
selected 1)~\emph{Wine Quality} dataset with the response variable \emph{quality}
2)~\emph{Housing} dataset with the response variable \emph{median value of owner-occupied homes}
and 3)~\emph{Auto MPG} dataset the response variable \emph{miles per gallon}.
In each of the dataset we scaled the response variable $y_r$ to the $[-1,+1]$ interval
and set the class label as $y := y_r \geq 0$. 
We randomly switched the class labels for 
three percent\footnote{We also performed the experiments with 1\% to 10\% of switched labels 
with the same trends.} of examples.
The true anomaly score was computed as the absolute difference 
between the original response variable $y_r$ and the (possibly switched) label.
Table~\ref{tab:uci} compares the agreement scores to the true score for all methods on 
(2/3, 1/3) train-test split. We see that SoftHAD either performed the best
or was close to the best method.

%\begin{figure}
%\begin{center}
%\includegraphics[width=\columnwidth, clip, viewport=25 256 581 530]{img/uci.pdf}
%\caption{Mean anomaly agreement score and variance (over 100 runs) for CAD methods on the 3 UCI ML datasets.}
%\label{fig:uci}%
%\end{center}
%\end{figure}
%\vskip -0.3cm
\begin{table}[htbp]  
\hskip -0.7cm	
 \begin{tabular}{r@{}c@{}|c|@{}c@{}}		
    %\addlinespace
    %\toprule		
        & \textbf{Wine Quality}\phantom{a}&\textbf{Housing}& \phantom{a}\textbf{Auto MPG} \\    				
	   \emph{QDA}& 75.1\% (1.3) & 56.7\% (1.5) & 65.9\% (2.9) \\
   \emph{SVM}  & 75.0\% (9.3) & 58.5\% (4.4) & 37.1\% (8.6) \\
    \emph{1-class SVM}  & 44.2\% (1.9) & 27.2\% (0.5) & 50.1\% (3.5) \\
    \emph{w$k$--NN}         & 67.6\% (1.4) & 44.4\% (2.0) & 61.4\% (2.3) \\
		\emph{SoftHAD}      & 74.5\% (1.5) & 71.3\% (3.2) & 72.6\% (1.7) \\	
    %\bottomrule		
    \end{tabular}%	
		\vskip 0.25cm
		\caption{Mean anomaly agreement score and variance (over 100 runs) for CAD methods on the 3 UCI ML datasets.}
  \label{tab:uci}%	
\end{table}%

\subsection{Medical data}
\label{sec:Dataset}
In this real-world experiment, we evaluated CAD on data extracted from
electronic health records (EHR) of 4,486 patients. Our goal was to detect 
unusual lab test orders or medication administrations.  
We divided EHRs into two groups:  a training set (2646 patients), and a test set (1840 patients).
For each patient, we segmented the data according to the length of the patient stay
where we considered all the patient data available at 8:00am each day.
%,
%which corresponds to the routine morning visit.
%Therefore, for a single patient, we obtained multiple patient-case instances.
%Therefore, we obtained multiple patient-case instances,
%as illustrated in Figure \ref{fig:pcp_segmentation}.
These patient instances were then converted into: (1) 9,282 features
%(Figure \ref{fig:pcp_features}  illustrates a subset of features
%generated for labs.),
and (2) 749 labels/tasks  --  reflecting whether a particular lab was ordered or a particular medication
was given within a 24-hour period.  This segmentation led to 51,492 patient-state instances,
such that 30,828 were used for training and 20,664 for testing.
More details can be found in \cite{hauskrecht2010conditional}.
\eat{For the evaluation we chose the 20 most frequent decisions for lab test orders
(such as GLU:Glucose, EOS:Eosinophil, K:Potassium,)
and 20 most frequent medication order decisions (such as Oxycodone, Potassium, Magnesium)
and 50 features associated for each of these.}

%\begin{center}
%\begin{figure}[t]
%\includegraphics[width=\columnwidth]{img/pcp_segmentation.png}
%\caption{Processing of data in the electronic health record: (1) segmentation of an EHR into multiple patient-state/decision instances, (2) transformation of these instances into a vector space representation of patient states and their follow-up decisions.}
%\label{fig:pcp_segmentation}%
%\end{figure}
%\end{center}

%\begin{figure}
%\begin{center}
%\includegraphics[width=\columnwidth, clip, viewport=220 230 414 329]{img/pcp_lab_features}
%\includegraphics[scale=0.25]{img/pcp_features.png}
%\caption{Examples of temporal features for continuous lab values: Last value: $A$,  Last value difference = $B-A$, Last \% change = $(B-A)/B$,  Last slope = $(B-A) / (t_B-t_A)$, Nadir = $D$, Nadir difference = $A-D$, Nadir \% difference = $(A-D)/D$}
%\label{fig:pcp_features}%
%\end{center}
%\end{figure}

\paragraph{Parameters for the graph-based algorithms}
\label{sec:Parameters}

To construct $G$, we computed the similarity weights as:
%\begin{equation}
$$w_{ij} = \exp\left[- \left(||\bx_i - \bx_j||_{2,\psi}^2 \right) / \sigma^2 \right],$$
%\label{eq:w}
%\end{equation}

\noindent where $\psi$ is a weighing of the features (we used Wilcoxon score \cite{hanley1982meaning}) and $\sigma$ is a length scale parameter.
%We weighted the features based on their discriminative power
%according to the univariate Wilcoxon score \cite{hanley1982meaning}.
%Since this score ranges from 0.5 to 1, we modify the score by subtracting 0.5
%and raising it to the power of 5 to make the differences between the weights larger.
%Including $p$ in the weight metric allows us to control the connectivity of the graph.
%Next, $\sigma$ is chosen so that the graph is reasonably sparse \cite{luxburg2007tutorial}.
We chose $\sigma$ as 10\% of the empirical variance of the Euclidean distances.
For each label, we sampled an equal number of positive and negative instances to construct a
$k$-NN graph. We set $k=75$, $c_l = 1$ and varied $\gamma_g$ and the graph size.

\paragraph{Scaling for multi-task anomaly detection}
\label{sec:Scaling}
%In the multi-task setting $\by = y^1y^2\dots y^l$.
%So far, we have described CAD only for a single task (anomaly in the single label).
In this dataset, we have 749 binary labels. We want to output an anomaly score which is comparable among the different tasks/labels
so we can, for example, set a unified threshold when the system is deployed in practice.
To achieve this score comparability, we propose a simple approach where we take the minimum and the maximum score obtained for the training set
and scale all scores for the same task linearly so that the score after the scaling ranges from 0 to 1.

\subsection{Results and Conclusion}
\label{sec:Results}
For the dataset described above, we computed the SoftHAD anomaly scores according to \eqref{eq:had}.
We asked the panel of 15 clinical experts to
evaluate the 222 patient case-label pairs (selected from $749\times20,664$ test case-label pairs),
such that every case-label was evaluated by 3 experts who decided whether the alert was clinically relevant.
We finally evaluated the performance of the CAD methods 
using the area under the ROC curve. 
%We compared our method to  SVM with RBF kernel (the runner up method from the synthetic experiment)
% 3) 1-class SVM with RBF kernel
%described in the beginning of this section.
\eat{
\begin{figure}
\begin{center}
\includegraphics[width=\columnwidth,clip, viewport=52 265 539 522]{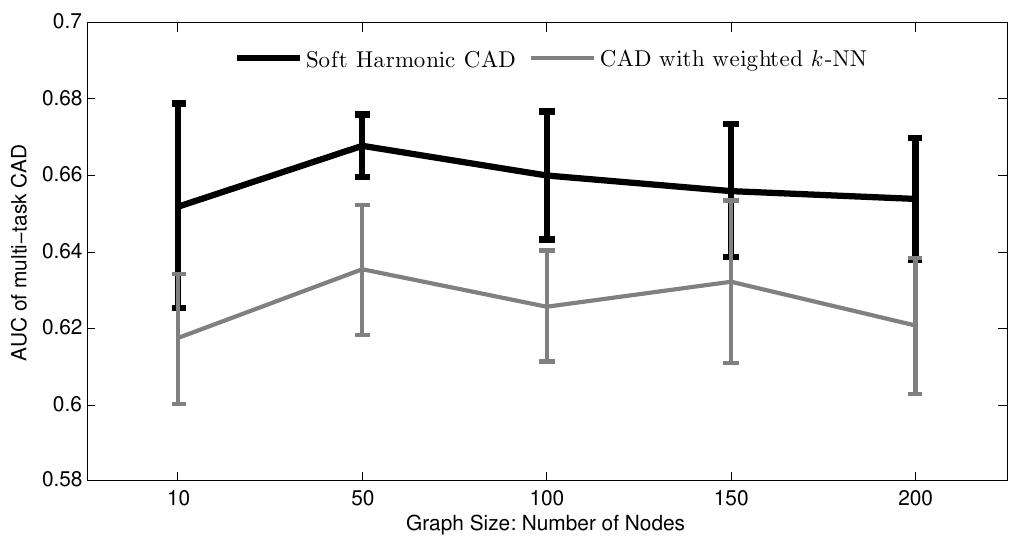}
\caption{Medical Dataset: Varying graph size. Comparison of 1) SoftHAD and 2) weighted $k$-NN on the same graph.}%
\label{fig:samplesize}%
\end{center}
\end{figure}
In Figure~\ref{fig:samplesize}, we fixed $\gamma_g = 1$ and varied the number of examples we sampled from the training set to construct the similarity graph and compared it to the weighted $k$--NN. The error bars show the variances over 10 runs.
Notice that the both of the methods are not too sensitive to the graph size.
This is due to the multiplicity adjustment  for the backbone graph (Section~\ref{sec:BackboneGraph}).
Since we used the same graph both for SoftHAD and weighted $k$--NN, we anticipate that we are able to outperform
weighted $k$--NN due to the label propagation over the data manifold and not only within the immediate neighborhood.
}

\begin{figure}
\begin{center}
\includegraphics[width=\columnwidth, clip, viewport=39 251 565 530]{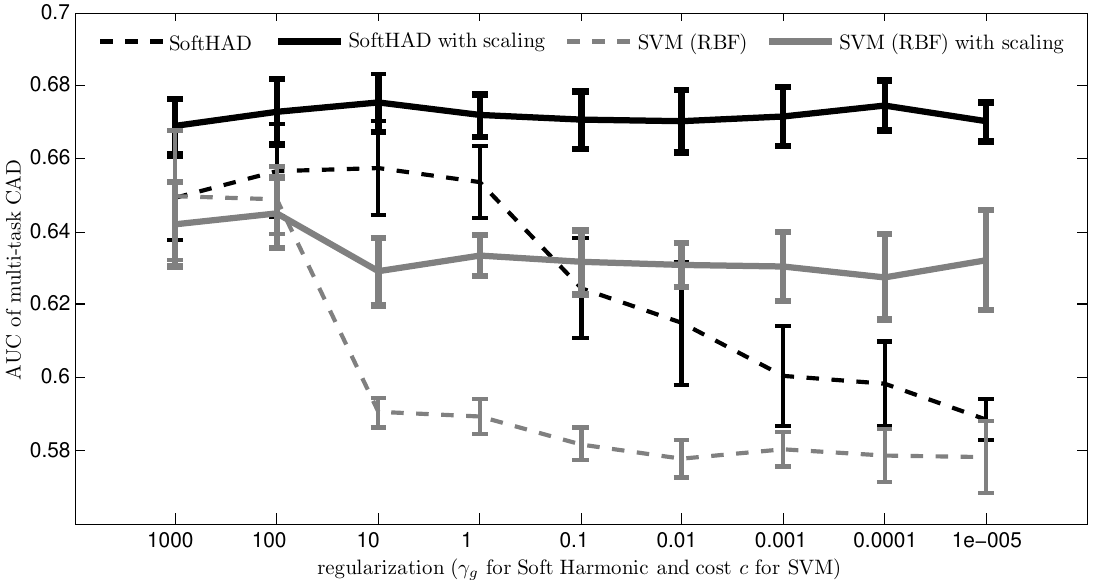}
\caption{Medical Dataset: Varying regularizer 1) $\gamma_g$ for SoftHAD 2) cost $c$ for SVM with RBF kernel.}
\label{fig:gg}%
\end{center}
\end{figure}

In Figure~\ref{fig:gg}, we compared SoftHAD vs.~CAD using SVM with RBF kernel for different regularization settings.
We sampled 200 examples to construct  $G$ (or train an SVM) and varied the $\gamma_g$ regularizer (or cost $c$ for SVM).
Scaling anomaly scores to the same range improved the performance of both methods and makes the methods less sensitive to the regularization settings.
We outperformed SVM approach over the range of regularizers.
In Figure~\ref{fig:samplesize}, we fixed $\gamma_g = 1$ and varied the number of examples we sampled from the training set to construct the similarity graph and compared it to the weighted $k$--NN. The error bars show the variances over 10 runs.
Notice that the both of the methods are not too sensitive to the graph size.
In future, we plan to extend the approach to the online anomaly detection.
This can be beneficial for the deployment of our SoftHAD method in hospitals.
%AUC for 1-class SVM with RBF was consistently below 55\% and we do not show it in the figure.
%We also compared the two methods with scaling adjustment for this multi-task problem (Figure~\ref{fig:gg}).

This research work was supported by grants 
R21LM009102, R01LM010019, and R01GM088224 
from the NIH. Its content is solely the responsibility 
of the authors and does not necessarily represent the 
official views of the NIH.

\begin{figure}
\begin{center}
\includegraphics[width=\columnwidth,clip, viewport=52 265 539 522]{img/figSIZE.pdf}
\caption{Medical Dataset: Varying graph size. Comparison of 1) SoftHAD and 2) weighted $k$-NN on the same graph.}%
\label{fig:samplesize}%
\end{center}
\end{figure}

%\bibliography{miki}
%\bibliographystyle{icml2011}

\end{document}